\documentclass[letterpaper, 10 pt, conference]{ieeeconf}  
\IEEEoverridecommandlockouts                              

\usepackage{stfloats}

\usepackage{pgfplots}
\DeclareUnicodeCharacter{2212}{−}
\usepgfplotslibrary{groupplots,dateplot}
\usetikzlibrary{patterns,shapes.arrows}
\pgfplotsset{compat=newest}
\usepackage{tikz}
\usetikzlibrary{shapes, arrows.meta, positioning}

\usepackage{cite}
\usepackage[normalem]{ulem}

\usepackage{times}
\usepackage{url}
\usepackage[colorlinks=True,allcolors=blue]{hyperref}
\usepackage[utf8]{inputenc}
\usepackage{graphicx}
\usepackage{amsmath}
\usepackage{amsfonts}
\usepackage{booktabs}
\usepackage{algorithm}
\usepackage{algorithmic}

\usepgfplotslibrary{groupplots}
\usepgfplotslibrary{fillbetween}
\usetikzlibrary{arrows,decorations.pathmorphing,positioning,fit,trees,shapes,shadows,automata,calc} 
\usetikzlibrary{patterns,arrows,arrows.meta,calc,shapes,shadows,decorations.pathmorphing,decorations.pathreplacing,automata,shapes.multipart,positioning,shapes.geometric,fit,circuits,trees,shapes.gates.logic.US,fit, matrix,arrows.meta, quotes}
\usetikzlibrary{backgrounds,scopes}

\usepackage{wrapfig}

\usepackage{mathtools}
\usepackage{accents}
\usepackage{mathrsfs}
\usepackage{mathtools}
\usepackage{amssymb}

\title{\LARGE \bf
Reference-Free Formula Drift with Reinforcement Learning:
From Driving Data to Tire Energy-Inspired, Real-World Policies
}

\author{
Franck Djeumou$^{1,2}$, Michael Thompson$^{1}$, Makoto Suminaka$^{1}$, and John Subosits$^{1}$
\thanks{$^{1}$
Toyota Research Institute, Los Altos, CA, USA}
\thanks{$^{2}$ 
MANE, Rensselaer Polytechnic Institute, Troy, NY, USA}
}

\begin{document}

\maketitle


\begin{figure*}[b]
    \centering
    \vspace*{-0.30cm}
    \includegraphics[width=\linewidth]{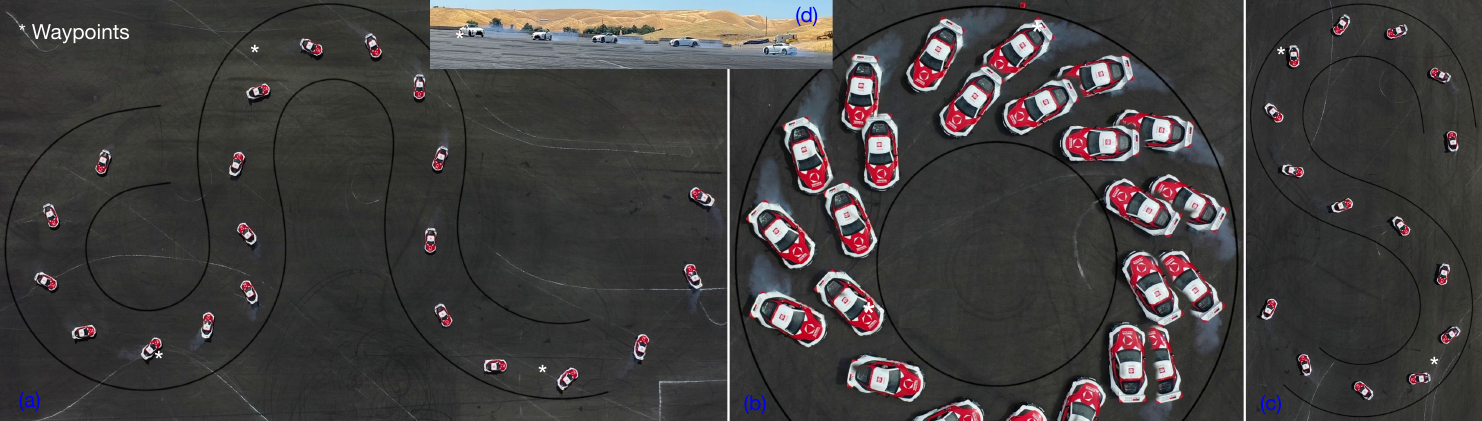}
    \vspace*{-0.5cm}
        \caption{Examples of RL policies smoothly pushing the Supra (a,b,c) and Lexus (d) to their limits of flexibility on various tracks and complex waypoint configurations, while enforcing pre-defined track bound constraints. The videos of the experiments can be found at \textcolor{blue}{\url{https://tinyurl.com/rl-drift}}. }
    \label{fig:intro}
\end{figure*}

\begin{abstract}
The skill to drift a car—i.e., operate in a state of controlled oversteer like professional drivers—could give future autonomous cars maximum flexibility when they need to retain control in adverse conditions or avoid collisions. We investigate real-time drifting strategies that put the car where needed while bypassing expensive trajectory optimization. To this end, we design a reinforcement learning agent that builds on the concept of tire energy absorption to autonomously drift through changing and complex waypoint configurations while safely staying within track bounds. We achieve zero-shot deployment on the car by training the agent in a simulation environment built on top of a neural stochastic differential equation vehicle model learned from pre-collected driving data. Experiments on a Toyota GR Supra and Lexus LC 500 show that the agent is capable of drifting smoothly through varying waypoint configurations with tracking error as low as $10$ cm while stably pushing the vehicles to sideslip angles of up to $63^\circ$.
\end{abstract}
\vspace*{-1mm}
\section{Introduction}\label{sec:intro}

Existing autonomous vehicles are constrained to operate in a conservative driving envelope with low lateral accelerations. 
However, in certain situations, it may be necessary to temporarily operate the vehicle beyond its natural stability limits to avoid a collision \cite{gray2012predictive,zhao2021collision, zhao2022justifying, li2023planning}. 
This style of driving is exemplified by drifting, a challenging cornering technique that involves deliberately saturating the rear tires to make the car slide while countersteering to maintain high sideslip angles.
Skilled human drivers display incredible vehicle control and agility in drifting competitions, routinely sliding their cars within inches of concrete walls \cite{FDjudging}.
Taking inspiration from their performance, this paper investigates an RL-based approach to stably push autonomous vehicles to their maximum agility potential.

Prior work has demonstrated steady-state stabilization \cite{Voser2010AnalysisAC, velenis2011steady, goh2016simultaneous}, course angle tracking \cite{werling2015robust}, and, ultimately, reference trajectory tracking \cite{goh2019automated} while drifting. 
Nonlinear model predictive control (MPC) with vehicle models based on expert knowledge \cite{gohAVEC2022, weber2023modeling, chen2023dynamic, goh2024beyond} or neural networks\cite{spielberg2021neural,djeumou2023autonomous,  ding2024drifting, djeumou2024diff, broadbent2024neural} has largely supplanted earlier explicit feedback schemes.
However, these MPC controllers rely on reference trajectories that are computed ahead of time by dynamic programming over a limited number of trajectory segments \cite{gray2012predictive}, rule and sampling-based planning \cite{zhang2018drift}, or trajectory optimization \cite{olofsson2013investigation}. 
With the sole exception of \cite{talbot2024optimal}, which demonstrates real-time trajectory replanning to a goal region, the question of how to drift with no reference trajectory while tracking arbitrary waypoint configurations remains largely unexplored.


Reinforcement learning (RL) could, in theory, provide drifting policies in such settings.
Cutler et al. show steady-state drift on a scale car with steering-only policies \cite{cutler2016autonomous}.
In \cite{orgovan2021autonomous} and \cite{cai2020high}, an RL agent learns to follow the trajectory of an expert human demonstration in simulation.
Validation of such approaches on full-size cars has been limited, although  T{\'o}th et al. demonstrate drift stabilization on a steady-state trajectory \cite{toth2024sim}. 
Most similar to our work, Domberg et al. \cite{domberg2022deep} design a reward for drifting along a general path that penalizes the lateral deviation and includes a heuristic term to encourage high, but bounded, sideslip angles.
Although the simulation results are promising, transfer to a scale car is less successful in scenarios with transient drift behavior. 
To date, no work has shown drifting policies tracking arbitrary waypoints on a full-scale vehicle.

We present the first RL-based drifting approach that 
\begin{itemize}
  \item builds on judging criteria of competitive drifting to achieve maximum agility via tire energy optimization.
  \item enables accurate tracking of varying and complex waypoint configurations without a reference trajectory.
  \item enables zero-shot transfer to the vehicle by training policies on physics-informed, neural stochastic differential equation (SDE) vehicle models from actual driving data.
\end{itemize}
We validate the proposed approach on a full-size Toyota GR Supra and Lexus LC 500. Our results show strong sim-to-real transfer capabilities, high agility, and high waypoint tracking accuracy on several tracks, including those in Figure \ref{fig:intro}.


\section{Data-Driven Simulation with Neural SDEs}
\label{sec:sde_model_sim}

This section introduces the uncertainty-aware and physics-constrained neural SDE vehicle model, trained from real-world data and used as a simulator for policy optimization.
The uncertainty-aware nature of the model obviates the need for explicit domain randomization for sim-to-real transfer.

\begin{wrapfigure}{l}{1.75in}
    \vspace*{-5mm}
    \includegraphics[width=\linewidth]{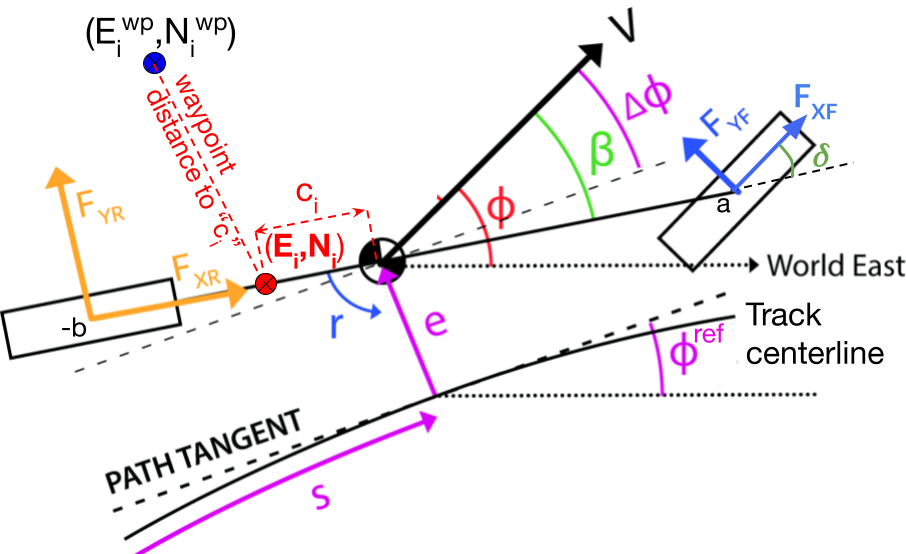}
    \vspace*{-6mm}
    \caption{Single-track model of a vehicle on a track with a single waypoint.}
    \vspace*{-3mm}
    \label{fig:bike-model-1}
\end{wrapfigure}
We employ the single-track assumption~\cite{Paden2016ASO,rajamani2011vehicle, Falcone2007PredictiveAS,Kong2015KinematicAD,Polack2017TheKB} to describe the vehicle's dynamics.
The vehicle position is expressed in a curvilinear coordinate system relative to a reference path ~\cite{Goh2019TowardAV,Goh2016SimultaneousSA,Subosits2021ImpactsOM}, as shown in Figure \ref{fig:bike-model-1}.
The position coordinate is given by the distance $s$ along the path, the relative heading $\Delta \phi$ to a planned course $\phi^{\mathrm{ref}}$, and the lateral deviation $e$ from the path centerline.
The vehicle state $x = [r, V, \beta, \omega_r, e, \Delta\phi, s]$ includes the yaw rate $r$, velocity $V$, sideslip angle $\beta$, rear wheelspeed $\omega_r$, lateral error $e$, and angular deviation $\Delta\phi$ while the control input $u = [\delta, \tau^{\mathrm{e}}]$ represents the desired steering angle and engine torque.

\subsection{Neural SDE vehicle model}
We assume access to a driving dataset $\mathcal{T} = (\tau_l)_l$ of vehicle trajectories $\tau = (\bar{x}_{t_j}, u_{t_j})_{j}$, where $\bar{x}_t = [r_t, V_t, \beta_t, (\omega_r)_t] \in \mathbb{R}^4$ and $u_t \in \mathcal{U} \subseteq \mathbb{R}^2$ are state and control estimates at time $t$. \emph{We emphasize that $\mathcal{T}$ contains regular driving and drifting attempts that are uncorrelated to the RL agent's task}

\textbf{Stochastic model.} We build on the neural SDE framework proposed in~\cite{djeumou2023learn} to learn expressive, uncertainty-aware vehicle dynamics models from $\mathcal{T}$. The model is given by
\begin{align}
	\renewcommand{\arraystretch}{1.3}
	\mathrm{d} 
	\begin{bmatrix}
	r \\
	V \\
	\beta \\
	\omega_r \\
	\end{bmatrix} = &
	\begin{bmatrix}
		\frac{a F^{\theta}_{yf} \cos (\delta) + a F^{\theta}_{xf} \sin(\delta) - b F^{\theta}_{yr}} {I^{\theta}_z} \\
		\frac{-F^{\theta}_{yf} \sin (\delta-\beta) + F^{\theta}_{xf} \cos (\delta-\beta)+F^{\theta}_{yr} \sin(\beta)+F^{\theta}_{xr} \cos (\beta)}{m} \\
		\frac{F^{\theta}_{yf} \cos (\delta-\beta)+F^{\theta}_{xf} \sin (\delta-\beta)+ F^{\theta}_{yr} \cos \beta - F^{\theta}_{xr} \sin \beta }{m V}-r \\
		\frac{G E^{\theta}(\tau^{\mathrm{e}}) -F^{\theta}_{xr}R}{I^{\theta}_w} \\
	\end{bmatrix} \nonumber \\
	&+ \Sigma^{\theta} ([r,V,\beta,\omega_r], u) \mathrm{d} W,
\end{align}
where $W$ is a $4$-dimensional Wiener process \cite{Stratonovich1966ANR, ito1951stochastic, kunita1997stochastic}, $\Sigma^{\theta} : \mathbb{R}^4 \times \mathcal{U} \rightarrow \mathbb{R}^{4 \times 4}$ is the SDE's diffusion term, and $a,b$, $m$, $R$, and $G$ represent the distance from the center of mass to the front axle, the distance to the rear axle, the vehicle mass, the wheel radius, and the gear ratio. Other parameters such as the yaw inertia $I^{\theta}_z$, drivetrain inertia $I^{\theta}_w$, or affine map $E^{\theta}$ from the engine to wheel torque are learned from $\mathcal{T}$.

\textbf{Tire forces. }To model the unknown forces $F^{\theta}_{yf}$, $F^{\theta}_{xf}$, $F^{\theta}_{yr}$, and $F^{\theta}_{xr}$, we build on the neural-$\mathrm{ExpTanh}$ tire model \cite{djeumou2023autonomous}, a physics-informed tire model that captures the nonlinearities and saturation effects of the road-tire interaction, and that has empirically shown to better predict tire forces than previous models in the literature. The forces are expressed as
\begin{align}
        &F^{\theta}_{xf} = 0, \: F^{\theta}_{yf} = \mathrm{ExpTanh}^{\theta}(\tan (\alpha_{f}); \: \mathrm{feat}_1), \label{eqn:tire-forces-1} \\
		& F^{\theta}_{\mathrm{tot}} = \mathrm{ExpTanh}^{\theta}(\tan^2 (\alpha_r) + f^{\theta}_0\sigma_r^2; \: \mathrm{feat}_2), \nonumber\\
        &\begin{bmatrix}
            F^{\theta}_{yr} \\
            F^{\theta}_{xr} \\
        \end{bmatrix} = \frac{\mathrm{NN}_0^{\theta}(\alpha_r, \sigma_r)}{\lVert \mathrm{NN}_0^{\theta}(\alpha_r, \sigma_r) \rVert}F^{\theta}_{\mathrm{tot}}, \: \sigma_r = \frac{R^{\theta} \omega_r -  V \cos \beta}{V \cos \beta},\nonumber\\
        &\alpha_f = \mathrm{atan} \frac{ V\sin \beta + a^{\theta} r}{V \cos \beta} - \delta, \: \alpha_r = \mathrm{atan} \frac{ V\sin \beta - b^{\theta} r}{V \cos \beta},\nonumber
\end{align}
where $\mathrm{ExpTanh}^\theta (z; \mathrm{feat}) \coloneqq f^\theta_1 +  f^\theta_2e^{-f^\theta_3|z|} \tanh \big( f^\theta_4(z-f^\theta_5) \big)$ is such that $(f_i^\theta)_{i=1}^5 = \mathrm{NN}^{\theta}(\mathrm{feat})$ satisfies $f^{\theta}_3,f^{\theta}_4 \geq 0$.
The variables $\mathrm{feat}_1 = [r, V,\beta]$ and $\mathrm{feat}_2 = [r, V,\beta, \omega_r]$ are features for $F^\theta_{yf}$ and $F^\theta_{\mathrm{tot}}$, while $f_0^{\theta}$ and $\mathrm{NN}_0^\theta$ approximate the coupling between the longitudinal and lateral tire dynamics. As in existing work, we have that $F_{xf} = 0$ since we assume a rear-wheel drive vehicle and no front brake application. 

\textbf{Neural SDE training.}
We compute the neural SDE's parameters as $\theta^{\mathrm{opt}} = \mathrm{argmin}_{\theta} \: \mathbb{E}_{\tau_{t:\mathrm{T}} \sim \mathcal{T}} \mathcal{J}^\theta_{\mathrm{nll}}(\tau_{t:\mathrm{T}})$, where $\tau_{t:\mathrm{T}}$ is a state-action sequence $(\bar{x}_{j}, u_{j})_{j=t}^{\mathrm{T}+t}$ of length $T$, and 
\begin{align}
	\mathcal{J}^\theta_{\mathrm{nll}}= \mathbb{E}_{\tilde{x}^\theta_{t:T}} \left[ \sum_{k=t}^{t+T} \lVert \bar{x}_{k} - \tilde{x}^\theta_{k} \rVert^2  _{(\Sigma^\theta_k)^{-1}}  + \log (\det (\Sigma^\theta_k )) \right], \label{eqn:loss-sde}
\end{align}
where $\Sigma^\theta_k \coloneqq \Sigma^{\theta}(\tilde{x}^{\theta}_k,u_k)$, the state sequence $\tilde{x}^{\theta}_{t:T}$ is a sample from a differentiable SDE integration scheme \cite{milstein1994numerical,kloeden2002numerical} with a fixed $\theta$, initial condition given by $\bar{x}_t$, and control sequence $u_{t:T}$, and $\| \cdot \|_A$ denotes the weighted norm by the matrix $A$.

\subsection{Track encoding and waypoint configuration}

In designing the simulation environment for the RL training, we encode all track information with the map $s \mapsto (E^{\mathrm{ref}}(s), N^{\mathrm{ref}}(s), \phi^{\mathrm{ref}}(s), \kappa^{\mathrm{ref}}(s), e_{\mathrm{min}}^{\mathrm{ref}}(s), e_{\mathrm{max}}^{\mathrm{ref}}(s))$, where
$E^{\mathrm{ref}}$, $N^{\mathrm{ref}}$, and $\phi^{\mathrm{ref}}$ are the east, north, and heading coordinates of the centerline, respectively, $\kappa^{\mathrm{ref}}$ is the track local curvature, and the lateral deviations $e_{\mathrm{min}}^{\mathrm{ref}}$ and $e_{\mathrm{max}}^{\mathrm{ref}}$ encode the track bounds.
Besides using $e$ to characterize the vehicle proximity to the boundaries, we also use the deviations $e^{\mathrm{front}}$ and $e^{\mathrm{rear}}$ of the front and rear bumpers given by
\begin{align}
	e^{\mathrm{front}} = e - a \sin (\beta - \Delta \phi), e^{\mathrm{rear}} = e + b \sin (\beta - \Delta \phi). \nonumber
\end{align}

\textbf{Path-dependent dynamics.} The remaining vehicle states $e$, $\Delta\phi$, and $s$ are well described by the equations of motion $\mathrm{d} e = V \sin (\Delta \phi) \mathrm{d} t$, $\mathrm{d} s = V \cos (\Delta \phi)(1 - e \kappa_{\mathrm{ref}}(s))^{-1} \mathrm{d} t$, and $\mathrm{d} (\Delta \phi) = (\dot{\beta} + r - \kappa^{\mathrm{ref}}(s) \dot{s}) \mathrm{d} t$. \emph{Therefore, with the learned $\theta^{\mathrm{opt}}$ and the path dynamics above, we can simulate the vehicle's motion on the track by integrating the neural SDE model defined on the full state $x = [r, V, \beta, \omega_r, e, \Delta\phi, s]$.}

\textbf{Waypoint setup.} Given the track information, one of the goals of the RL agent is to minimize the distance to a set of waypoints $\{ (c_i, s^{\mathrm{wp}}_i, E^{\mathrm{wp}}_i, N^{\mathrm{wp}}_i) \}_{i=1}^{\mathrm{n}^{\mathrm{wp}}}$, where $c_i \in [-b, a]$ (illustrated in Figure \ref{fig:bike-model-1}) parameterizes a point on the vehicle's longitudinal centerline whose planar coordinates must coincide with the east $E^{\mathrm{wp}}_i$ and north $N^{\mathrm{wp}}_i$ coordinates at $s=s^{\mathrm{wp}}_i$. \emph{For example, if $c_i = 0$, $c_i = -b$, or $c_i = a$, the waypoint must coincide with the center of the vehicle, the front bumper, or the rear bumper, respectively, at $s=s^{\mathrm{wp}}_i$.} Through elementary kinematics, we can express the east $E_i$ and north $N_i$ coordinates of $c_i$ at the path distance $s$ as
\begin{align}
	& E_i(s) = E^{\mathrm{ref}}(s) - e \sin (\phi^{\mathrm{ref}}(s)) - c_i \sin (\Phi(s)), \\
	& N_i(s) = N^{\mathrm{ref}}(s) - e \cos (\phi^{\mathrm{ref}}(s)) + c_i \cos (\Phi(s)), 
\end{align}
where $\Phi(s) = \phi^{\mathrm{ref}}(s) + \Delta \phi - \beta$.
\section{Tire Energy-Inspired RL Drifting Policies}

In this section, we describe the design of an RL agent that learns to drift autonomously by maximizing tire energy absorption while accurately tracking arbitrary waypoint configurations and safely staying within the track bounds. We train the agent in a simulation environment built on top of the learned neural SDE vehicle model described in Section~\ref{sec:sde_model_sim}.

\subsection{Observation space and policy design}

We design the observations and actions of the RL agent to capture essential features of the environment and vehicle's state. Such features enforce generalization to unseen tracks and waypoints as well as smooth and human-like control.

\textbf{Observation space.}
The agent's observations are all derived from the vehicle's state $x$ and past applied control.
\begin{itemize}
    \item \textit{State variables.} We use only the components $r$, $V$, $\beta$, $\omega_r$, and $\Delta \Phi$ of $x$ in the observation design.
    \item \textit{Path curvature gradient.} Given a lookahead $\mathrm{d}^{\mathrm{look}}$ and horizon $\mathrm{n}^{\mathrm{look}}$, we compute $\kappa_i = \kappa^{\mathrm{ref}}(s + \mathrm{d}^{\mathrm{look}} i )$ for $i \leq \mathrm{n}^{\mathrm{look}}$.
    Then, the observation is given by the gradient vector $\Delta \kappa = \{ \kappa_{i} - \kappa_{i-1} \}_{i=1}^{\mathrm{n}^{\mathrm{look}}}$, a design choice to improve the generalization to changes in the curvature.
    \item \textit{Waypoint distance.} The corresponding observations are quantities depending on $\Delta s_i = s^{\mathrm{wp}}_i - s$ and the $2$-norm distance $d_i$ between $(E^{\mathrm{wp}}_i, N^{\mathrm{wp}}_i)$ and $(E_i(s)$, $N_i(s))$:
    \begin{align}
        & \Delta s^{\mathrm{wp}}_i = 1 - \exp \{ - \gamma_1 \max(0, \Delta s_i) \}, \text{ } \\
        & d^{\mathrm{wp}}_i = \mathcal{B}(\Delta s_i) \cdot \frac{1}{d_i + 10^{-6}}, \text{ where}\\
        & \mathcal{B}(z) = \begin{cases}
            \exp \Big\{ \frac{\gamma_2}{ (z + \gamma_3)^2 - \gamma_4^2 } \Big\}, & \text{if } |z + \gamma_3| < \gamma_4, \\
            0, & \text{otherwise},
        \end{cases}
    \end{align}
    with $\mathcal{B}$ being a smooth function that peaks at $z = - \gamma_3$, is nonzero for $z$ in the interval $(-\gamma_3 - \gamma_4, -\gamma_3 + \gamma_4)$, and zero outside. The parameter $\gamma_2$ controls how fast the function reaches its maximum value.  
    \emph{Intuitively, $\mathcal{B}(\Delta s_i)$ acts as a masking function enabling the agent to focus only on waypoints that are within a certain range of the vehicle's path distance, while $\Delta s^{\mathrm{wp}}_i$ informs both on the proximity and whether the waypoint has been traversed or not.} We then use both $\Delta s^{\mathrm{wp}} = \{ \Delta s^{\mathrm{wp}}_i \}_{i=1}^{\mathrm{n}^{\mathrm{wp}}}$ and $d^{\mathrm{wp}} = \{ d^{\mathrm{wp}}_i \}_{i=1}^{\mathrm{n}^{\mathrm{wp}}}$ as waypoint observations, or only the next few closest waypoints if $n^{\mathrm{wp}}$ is too high.
    \item \textit{Proximity to track boundaries.} We use as observation $\Delta e^{\square} = \min (e^{\mathrm{ref}}_{\mathrm{max}}(s) - e^{\square}, e^{\square} -e^{\mathrm{ref}}_{\mathrm{min}}(s) ) $, where $e^{\square}$ is either $e^{\mathrm{front}}$ or $e^{\mathrm{rear}}$.
    \item \textit{Past control and time step.} We include the past control $u_{-1}$ and the duration $\Delta t$ over which it was applied.
\end{itemize}


\textbf{Policy design.}
We modify the policy network structure to ensure that it outputs physically-feasible control inputs in $\mathcal{U}$ and such that the rate of change of the inputs is bounded by $\mathcal{U}^{\mathrm{rate}}$. Given an observation $o$, we define the policy as 
\begin{align}
    \pi^{\theta}(o) = u_{-1} + \Delta t \big( \hat{\mathcal{U}}^\mathrm{rate} + \tilde{\mathcal{U}}^{\mathrm{rate}} \tanh \big(\bar{\pi}^{\theta}(\bar{o}) \big) \big), \label{eqn:policy-design}
\end{align}
where $\hat{\mathcal{U}}^{\mathrm{rate}}$ and $\tilde{\mathcal{U}}^{\mathrm{rate}}$ are adequate constants to transform $[-1,1]\times[-1,1]$ back to $\mathcal{U}^{\mathrm{rate}}$,
the policy $\bar{\pi}^{\theta}$ is a stochastic Gaussian policy network, and $\bar{o}$ is the observation $o$ without the time step. Finally, the environment clips $\pi^{\theta}$ in $\mathcal{U}$ before simulating the neural SDE vehicle model.

\subsection{Reward design}
Our reward function is inspired by judging criteria for single-car runs in professional drift competitions. The drivers are evaluated based on the extent to which they can maintain ``a high degree of angle,'' ``maintain momentum,'' and drive with ``fluidity,'' all while accurately hitting a small number of pre-defined points on the track \cite{FDjudging}.
We propose tire energy as the reward signal to enable drifting with ``high angles'' without relying on a hard-coded heuristic on the slip angle. 
We then add rewards on the waypoints, track progress, smooth actions and drift transitions, and track bounds.

\textbf{Tire energy.} The corresponding reward is given by
\begin{align}
    \mathrm{R}^{\mathrm{tire}} = \Delta t \big( &| ( V \sin (\beta) + b r ) F_{yr} | \nonumber\\
    &+ | (  R \omega_{r} - V \cos (\beta) \big) F_{xr}  | \big), \label{eqn:tire-en}
\end{align}
where the first term represents the energy contribution from the rear lateral tire force and corresponding velocity slip, and the second term is the longitudinal contribution.
This term strongly encourages the vehicle to maintain a large drift angle without directly rewarding the sideslip angle, which could interfere with behavior during transient drift maneuvers where the slip angle must pass through zero.

\textbf{Waypoints.} We use separate rewards for task progress $\Delta s^{\mathrm{wp}}$  and the distance to waypoints. They are given by
\begin{align}
    \mathrm{R}^{\mathrm{pgr}} &= \sum\nolimits_{i=1}^{\mathrm{n}^{\mathrm{wp}}} (1-\Delta s^{\mathrm{wp}}_i) \text{ and } \mathrm{R}^{\mathrm{wp}} = \sum\nolimits_{i=1}^{\mathrm{n}^{\mathrm{wp}}} d^{\mathrm{wp}}_i.
\end{align}

\textbf{Smoothness.} We enforce smooth transitions and human-like control inputs during drift maneuvers by penalizing large accelerations of the vehicle and the rate of changes of control inputs. The corresponding rewards are given by
\begin{align}
    \mathrm{R}^{\mathrm{eqbr}} &= - (\lambda_r \dot{r}^2 + \lambda_\beta \dot{\beta}^2 + \lambda_\omega \dot{\omega}_r^2 + \lambda_V \dot{V}^2), \text{ and} \label{eqn:low-acc} \\
    \mathrm{R}^{\mathrm{rate}} &= - \Delta t^{-2} \lambda_{\mathrm{rate}} \cdot (u - u_{-1})^2, \label{eqn:control-rate}
\end{align}
where $\lambda_r, \lambda_\beta, \lambda_\omega, \lambda_V \in \mathbb{R}_+$ and $\lambda_{\mathrm{rate}} \in \mathbb{R}^2_+$ are hyperparameters to weight the contribution of each term.

\textbf{Track bounds.} We penalize proximity to track edges with
\begin{align}
    \mathrm{R}^{\mathrm{edge}} = - \sum\nolimits_{\square \in \{ \mathrm{front}, \mathrm{rear} \}} \max (10^{-6}, \Delta e^{\square})^{-1}, \label{eqn:safe-rew}
\end{align}
where the reward is inversely proportional to the bumper's distance to the edges and reaches $-10^6$ beyond the edges.

The final reward is given as a linear combination of the above rewards weighted by hyperparameters $\lambda_{\mathrm{tire}}$, $\lambda_{\mathrm{pgr}}$, $\lambda_{\mathrm{wp}}$, $\lambda_{\mathrm{edge}}$, and weighted by $1$ for the rewards $\mathrm{R}^{\mathrm{eqbr}}$ and $\mathrm{R}^{\mathrm{rate}}$.

\subsection{Massively parallel RL agent training}
Although any adequate off-the-shelf RL algorithm can be used to train the drifting policy, we choose the Proximal Policy Optimization (PPO) algorithm \cite{schulman2017proximal} due to its simplicity, easy customization, and massive parallelization capabilities.

\textbf{Episode termination.} During policy training and evaluation, we terminate an episode when any of the lateral deviations $e$, $e^{\mathrm{front}}$, or $e^{\mathrm{rear}}$ exceeds the track bounds $e^{\mathrm{ref}}_{\mathrm{max}}(s)$ or $e^{\mathrm{ref}}_{\mathrm{min}}(s)$. 
The episode also terminates for non-increasing path distance evolution over a single environment step, which indicates the vehicle is going backward. Finally, we limit the episode duration with a maximum path distance $s^{\mathrm{goal}}$.

\textbf{Policy smoothness and noise robustness.}
Existing approaches \cite{cai2020high,domberg2022deep,toth2024sim} and their results have highlighted the challenges of obtaining smooth, non-oscillatory, and human-like control from an RL drift policy, even for steady-state drift.
In addition to the reward $\mathrm{R}^{\mathrm{rate}}$ that encourages smooth policies by penalizing large changes in the controls, we add to the PPO's policy loss an observation-dependent term $\mathcal{J}^{\theta}_{\mathrm{jac}}$, weighted by $\lambda_{\mathrm{jac}}$. The term $\mathcal{J}^{\theta}_{\mathrm{jac}}$ enforces directly in the network structure smoothness, diminish oscillatory response, and increase noise robustness.
To this end, we regularize an approximation of the Frobenius norm of the policy network's Jacobian \cite{hoffman2019robust}. 
This approach has not only been shown to improve generalization capabilities and noise robustness when compared to other existing regularization techniques but also adds only a small computational overhead to the RL training algorithm. $\mathcal{J}^{\theta}_{\mathrm{jac}}$ is given by
\begin{align}
    \mathcal{J}^{\theta}_{\mathrm{jac}}(o) = \mathbb{E}_{\nu \in \mathcal{N}(0, \mathbb{I})} \big[ \| \nabla_{\bar{o}} \{ \bar{\pi}_{\mathrm{mean}}^{\theta}(\bar{o}) \cdot ( \nu \| \nu \|^{-1}) \} \|^2 \big], \label{eqn:jac-loss}
\end{align}
where $\bar{\pi}_{\mathrm{mean}}^{\theta}$ represents the mean of the Gaussian network $\bar{\pi}^{\theta}$ in \eqref{eqn:policy-design} and $\nu \in \mathbb{R}^2$ is a vector sampled from the normal distribution $\mathcal{N}(0, \mathbb{I})$. The loss $\mathcal{J}^{\theta}_{\mathrm{jac}}$ can be added to the PPO-Clip objective when performing mini-batch updates.

\textbf{Parallel training.}
We vectorize the integration scheme of the neural SDE model to enable massively parallel training. Each initial state is sampled from a distribution that covers a wide range of initial conditions and with the path distance randomly sampled to satisfy $s \leq s^{\mathrm{goal}}$. To avoid the policy overfitting a fixed time step, we randomly sample the integration time step $\Delta t$ from a uniform distribution.

\section{Experimental Results}

We validate our approach on two vehicles with different physical properties and control capabilities. First, we demonstrate zero-shot transfer capabilities to full-size vehicles from a small amount of driving data. Then, we verify that the learned policies can generalize to varying waypoint configurations, and are robust to changes in the environment. Finally, we conduct an ablation study to show the benefits of Jacobian regularization for smooth, human-like control responses. Unless explicitly stated otherwise, all variables are in the SI system of units.

\subsection{Experimental platforms and  training setup}

\begin{figure*}[t]
    \centering
    \includegraphics[width=\linewidth]{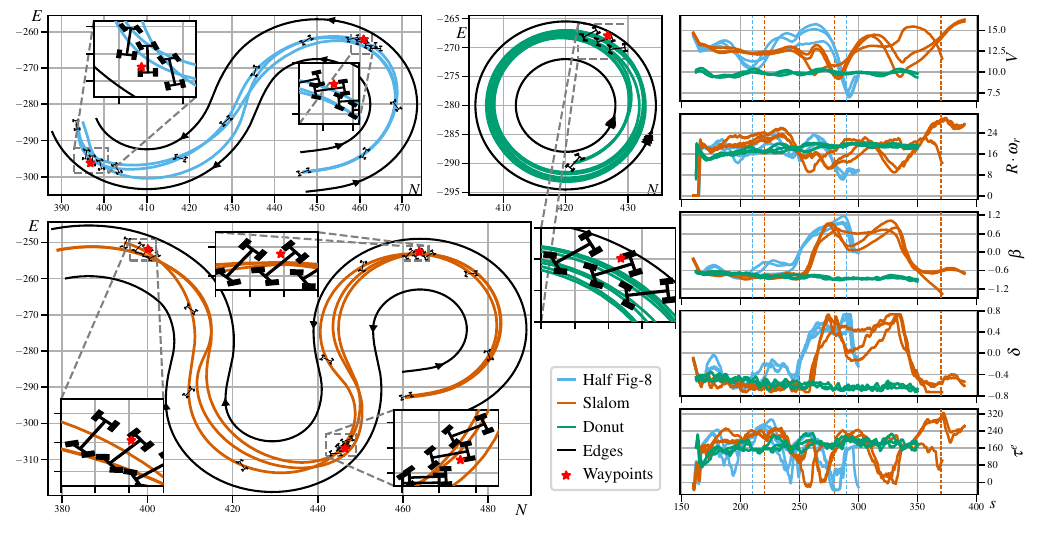}
    \vspace*{-0.8cm}
    \caption{Drifting with the Supra: the best runs per policy average $21$ cm waypoint tracking error while reaching slip angles as high as $63^\circ$.}
    \label{fig:keisuke-results}
    \vspace*{-5mm}
\end{figure*}
We deploy the approach on a Toyota GR Supra and a Lexus LC 500, as shown in Figure~\ref{fig:intro}. 
The Supra is modified with a more powerful engine, drift-specific steering and suspension, and more responsive actuators, while the Lexus is merely modified to permit autonomous control, making it a particularly challenging platform for autonomous drifting.
The vehicles' large differences in dynamics response make them ideal platforms for evaluating RL policies and their ability to push the limits of the vehicle's performance.
For both vehicles, we use onboard vehicle state estimation using a GPS and IMU, as well as the CPU of an onboard ruggedized PC to evaluate the policy network and send the steering and engine torque commands to a hard real-time computer (dSpace MicroAutoBox II) for low-level control. We refer to \cite{djeumou2024diff} and \cite{ding2024drifting} for details on the Supra and Lexus, respectively, and their parameters $a$, $b$, $R$, $m$, $\mathcal{U}$, and $\mathcal{U}^{\mathrm{rate}}$.

\textbf{Driving dataset.}
We train neural SDE-based simulators for the Supra and Lexus using manual and autonomous drifting trajectories collected from the vehicles. The Supra dataset consists of $17$ attempts at drifting and tracking the centerline of a donut and a figure-8 trajectory, while the Lexus contains $21$ attempts, each attempt lasting between $10$ and $60$ seconds. 
Besides, we collect the Lexus dataset by alternating between two set of tires with different characteristics: $14$ trajectories are from tire type $1$ ($T_1$) and the rest from tire type $2$ ($T_2$).

\textbf{Training neural SDE-based simulators.}
We use the same architecture for each vehicle's neural SDE-based simulator. The Supra's simulator is trained exclusively on the Supra dataset, and the same is true for the Lexus. $\Sigma^\theta$ is a diagonal matrix parametrized by a neural network with two hidden layers of $16$ neurons each and $\tanh$ activations. Following the architecture of $\Sigma^\theta$, $\mathrm{NN}_0^{\theta}$ and $\mathrm{NN}^{\theta}$ in the tire force design only have $8$ neurons each.
The unknown parameters $I^{\theta}_z$, $I^\theta_{w}$, and $f_0^{\theta}$ are learnable scalar values. 
We train the models using Adam optimizer \cite{kingma2017adam} with a learning rate of $10^{-3}$ and a batch size of $128$. Each state-action sequence in a batch of data is randomly sampled such that its duration is between $0.5$ and $2$ seconds. We use the Euler-Maruyama integration scheme to compute the loss $\mathcal{J}^\theta_{\mathrm{nll}}$ \eqref{eqn:loss-sde}, and we use $5$ predicted particles to estimate the expectation in the loss.

\textbf{RL agent design and hyperparameters.}
The agent can observe changes $\Delta \kappa$ of the track curvature up to $40$ meters ahead: we use $n^{\mathrm{look}} = 4$ and $d^{\mathrm{look}} = 10$. We use $\gamma_1 = 0.01$, $\gamma_2 = 20$, $\gamma_3 = -24$, and $\gamma_4 = 26$ to define the waypoint observations $\Delta s^{\mathrm{wp}}$ and $d^{\mathrm{wp}}$. That is, the observation $d_i^{\mathrm{wp}}$ of the $i$-th waypoint is zero except when $s - s_i^{\mathrm{wp}} \in [-48, 2]$. The reward weights are given by $\lambda_{\mathrm{tire}} = 2.5$, $\lambda_{\mathrm{pgr}} = 10$, $\lambda_{\mathrm{wp}} = 5000$, $\lambda_{\mathrm{edge}}= 0.001$, $\lambda_{\mathrm{r}} = \lambda_{\beta} = 1.0$, $\lambda_{\omega} = 0.01$, $\lambda_V = 0$, and $\lambda_{\mathrm{rate}} = [0.1, 10^{-7}]$. 
We randomly sample the time step $\Delta t$ between $0.01$ and $0.05$ seconds for each environment step.
We use $\lambda_{\mathrm{jac}} = 10^{-5}$ for the Jacobian regularization term and use a single sample $\nu$ to estimate the expectation in $\mathcal{J}_{\mathrm{jac}}^\theta$ \eqref{eqn:jac-loss}.
We use the default hyperparameters of PPO, except for choosing $0.0001$ as the learning rate and vectorizing the training over $2048$ environments. 


\subsection{Zero-shot, sim-to-real transfer across different vehicles}

We evaluate the proposed approach on a range of pre-defined track and waypoint configurations. The experiments demonstrate that, given a vehicle, track, and waypoint configuration, the RL policy trained on the corresponding neural SDE simulator can (a) be directly deployed on the actual vehicle and (b) achieve high performance in terms of agility (sideslip angle) and waypoint accuracy (see Table \textcolor{blue}{1}), all while being stable and driving within the track bounds.

\begin{table}[!hbt]
    \centering
    \vspace*{-2mm}
    \textbf{Table 1: Proximity error to the waypoints.}
    \vspace*{1mm}
    \label{tab:supra_all}
    \begin{tabular}{ |c|c|c|c|c|c|c|  } 
          \hline
          & \multicolumn{1}{|c}{Donut} & \multicolumn{2}{|c|}{Figure-8} & \multicolumn{3}{|c|}{Slalom}\\
         Dist (cm) & $\mathrm{wp}_1$ & $\mathrm{wp}_1$ & $\mathrm{wp}_2$ & $\mathrm{wp}_1$ & $\mathrm{wp}_2$ & $\mathrm{wp}_3$ \\ 
          \hline
         $\mathrm{Supra}$ & $17.2$ &  $34.1$ & $44.3$ & $27.4$ & $37.3$ & $48.8$\\
         \hline
         $\mathrm{Lexus}$ & $9.61$ & $32.2$ & $16.4$ & - & - & -\\ 
         \hline
    \end{tabular}
    \vspace*{-2mm}
\end{table}

\textbf{Drifting with the Supra.}
We consider three track configurations: a donut, half a figure-8, and slalom. We pick the width $e^{\mathrm{ref}}_{\mathrm{max}} - e^{\mathrm{ref}}_{\mathrm{min}}$ of the track to be no more than $14$ meters (around $3 \times$ the Supra's length). We investigate scenarios with $1$ to $3$ rear bumper-based waypoints ($c_i = -b$, the rear bumper must hit the waypoint), where some of them are as close as $0.5$ meters away from the track edges. Figure \ref{fig:keisuke-results} shows multiple runs of each track-and-waypoint-specific RL policy on the Supra.
We observe that the policies achieve high waypoint proximity with average and standard deviation values of $17.2 \pm 8$ cm, $39.2 \pm 16$ cm, and $37.8 \pm 27$ cm on the donut, figure-8, and slalom-based tracks, respectively.
Additionally, the learned policies can maintain stable drift at high sideslip angles with the vehicle operating close to its steering limits. For example, on the donut track, the vehicle gradually increases its slip angle until reaching $-55^\circ$ (and the steering angle limit of $-0.75$ rad, see Figure \ref{fig:keisuke-results}). This is an additional sideslip of $-15^\circ$ compared to reported results from reference trajectory and model-based control solutions \cite{djeumou2024diff} on the same track but without waypoint constraints. A similar trend can be observed for the figure-8 and slalom maps when compared to reported results in the literature \cite{djeumou2024diff}.
Interestingly, while the trajectories vary across each run of a given policy, in all cases, the vehicle stays within the road edges and closely approaches the waypoints.

\textbf{Drifting with the Lexus.}
Contrary to the Supra, it is challenging to use the Lexus for aggressive drift maneuvers without spinning out due to limitations in its actuator response.
We train and evaluate RL policies on a few modifications of the Supra's track and waypoint configurations. In particular, we modify the size of the tracks and include both front and rear bumper-based waypoints.
We report our findings in Figure \ref{fig:leia-results}. We observe the same trend as with the Supra, where the Lexus achieves an average waypoint tracking error of $9.61 \pm 3$ cm and $24.3 \pm 11$ cm on the donut and figure-8 (Fig8, $T_{1-2}-T_1$), respectively, while operating at the limits of its steering capabilities to reach slip angles as high as $45^\circ$.

\begin{figure}[!hbt]
    \centering
    \vspace*{-2mm}
    \includegraphics[width=0.8\linewidth]{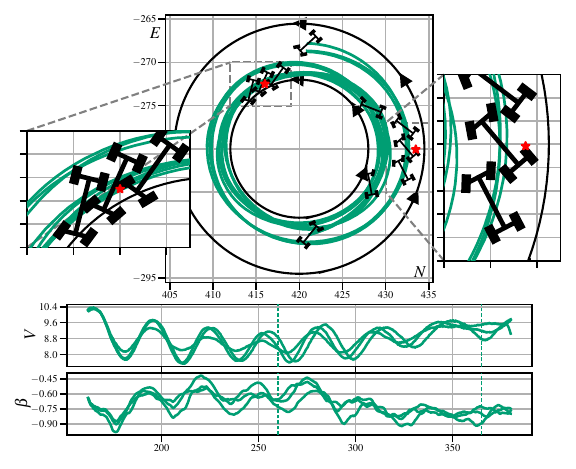}
    \vspace*{-0.4cm}
    \caption{The RL policy shows generalization to new waypoints on the Supra.}
    \label{fig:keisuke-wps-gen}
    \vspace*{-4mm}
\end{figure}
\subsection{Policy generalization and robustness}
\textbf{Generalization to waypoint configurations.}
Instead of optimizing the policy for a fixed waypoint configuration, we modify the simulator to initialize each episode with a random waypoint configuration and train the RL policy to adapt and generalize to changing test conditions. The waypoint selection process randomly picks front or rear bumper-based waypoints, with at least $50$ meters between them and as close as $0.5$ meters away from both track edges. After the policy is trained, we fix the waypoint configuration as illustrated in Figure \ref{fig:keisuke-wps-gen} and evaluate three runs of the policy on the Toyota Supra. Figure \ref{fig:keisuke-wps-gen} shows that the RL policy can perform well on a new waypoint configuration, with an average tracking error of $27 \pm 4$ cm for the first (front bumper) waypoint and $17.9 \pm 3$ cm for the second (rear bumper) waypoint. Figure \ref{fig:keisuke-wps-gen} also illustrates how smooth the RL policy is at transiting between the inside and outside waypoints while maintaining high slip angles.
Additionally, we evaluate the policy in simulation with $20$ different waypoint configurations and obtain an average tracking error of $23.3 \pm 6$ cm.

\begin{figure*}[t]
    \centering
    \includegraphics[width=\linewidth]{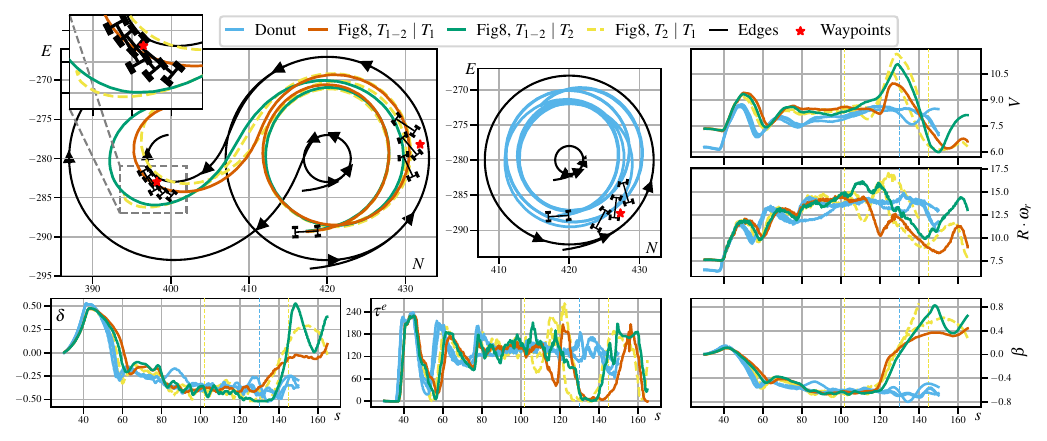}
    \vspace*{-0.8cm}
    \caption{Drifting with the Lexus: the best runs per policy average $11$ cm waypoint tracking error while reaching slip angles as high as $45^\circ$.}
    \label{fig:leia-results}
    \vspace*{-5mm}
\end{figure*}

\textbf{Robustness to environmental changes.}
We evaluate the robustness of the learned policies by swapping the Lexus tires between type $T_1$ and $T_2$, and analyzing the drift performance. The figure-8 experiment shown in Figure \ref{fig:leia-results} summarizes our findings. We show that the policy (\emph{Fig8, $T_2$})--trained on a neural SDE simulator fitting the Lexus dataset with tire $T_2$ only--when evaluated on the Lexus with tire $T_1$ (labeled as \emph{Fig8, $T_2 | T_1$} in the plot), achieves low waypoint tracking accuracy while often spinning out or going off track.
In contrast, the policy (\emph{Fig8, $T_{1-2}$})--trained on a simulator fitting both tires $T_1$ and $T_2$--when evaluated on $T_1$ and $T_2$ tires, can drift smoothly without spinning out or going off track. We observe better waypoint tracking accuracy when, at test time, the Lexus is equipped with tire $T_1$ (due to the ratio of $T_1$ trajectories in the dataset) compared to a more conservative driving and high waypoint tracking error when the Lexus is equipped with $T_2$. Such results suggest that model randomization during training can be used with the proposed framework to improve policy generalization.

\subsection{Ablation study: Jacobian regularization}
\begin{figure}[!hbt]
    \centering
    \vspace*{-6mm}
    \includegraphics[width=0.9\linewidth]{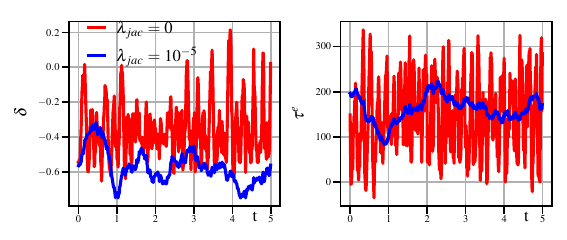}
    \vspace*{-0.5cm}
    \caption{The Jacobian regularization yields smooth and noise-robust policies.}
    \label{fig:jac-reg}
    \vspace*{-2mm}
\end{figure}

We investigate the effect of $\lambda_{\mathrm{jac}}$ on the RL policy's control performance. 
We train in an identical manner two policies in simulation on the Supra donut track: One without Jacobian regularization $\lambda_{\mathrm{jac}} = 0$ and one with $\lambda_{\mathrm{jac}} = 10^{-5}$. The resulting policies yield similar performance in terms of the total reward attained, and Figure \ref{fig:jac-reg} shows the steering and engine torque response on a five-second snapshot when simulating the policies.
The figure shows that the policy trained with Jacobian regularization achieves a smoother control response, lower oscillations, and higher robustness to noise, making it better suited to deployment on hardware. 
\section{Conclusion}

We propose the first RL-based drifting approach applied to a full-size vehicle that can drift, without a reference trajectory, across a general path defined by waypoints, all while pushing the car to its limits of agility through the principle of maximum tire energy absorption.
Extensive experiments with a Toyota GR Supra and Lexus LC 500 demonstrate the effectiveness of the approach.

Although we show promising results in using RL to navigate through complex waypoint configurations with high agility, the current formulation, similar to existing approaches, depends on full state estimation and precise track information. 
Future research could investigate drifting policies that work by fusing partial state estimates and visual-based measurements from LiDAR or RGB-D cameras.
Other interesting future research directions are policies that leverage brake actuators to improve stability and flexibility, adapt to changes in road conditions, and generalize across different platforms or tracks.


\clearpage
\newpage

\bibliographystyle{ieeetr}
\bibliography{bibliography}
\end{document}